\documentclass[sigconf]{acmart}
\pdfoutput=1
\usepackage{algorithm}
\usepackage{algorithmic}
\AtBeginDocument{%
  \providecommand\BibTeX{{%
    \normalfont B\kern-0.5em{\scshape i\kern-0.25em b}\kern-0.8em\TeX}}}

\copyrightyear{2022}
\acmYear{2022}
\setcopyright{acmlicensed}\acmConference[ICONS 2022]{International Conference on Neuromorphic Systems}{July 27--29, 2022}{Knoxville, TN, USA}
\acmBooktitle{International Conference on Neuromorphic Systems (ICONS 2022), July 27--29, 2022, Knoxville, TN, USA}
\acmPrice{15.00}
\acmDOI{10.1145/3546790.3546803}
\acmISBN{978-1-4503-9789-6/22/07}

\begin{document}

\title{Efficient Spike Encoding Algorithms for Neuromorphic Speech Recognition}

\author{Sidi Yaya Arnaud Yarga}
\email{sidi.yaya.arnaud.yarga@usherbrooke.ca}
\affiliation{
  \institution{Department of Electrical and Computer Engineering\\Université de Sherbrooke}
  \city{Sherbrooke}
  \state{QC}
  \country{Canada}
}
\author{Jean Rouat}
\email{jean.rouat@usherbrooke.ca}
\affiliation{
  \institution{Department of Electrical and Computer Engineering\\Université de Sherbrooke}
  \city{Sherbrooke}
  \state{QC}
  \country{Canada}
}
\author{Sean U. N. Wood}
\email{sean.wood@usherbrooke.ca}
\affiliation{
  \institution{Department of Electrical and Computer Engineering\\Université de Sherbrooke}
  \city{Sherbrooke}
  \state{QC}
  \country{Canada}
}

\renewcommand{\shortauthors}{Yarga, Rouat and Wood}

\begin{abstract}
Spiking Neural Networks are known to be very effective for neuromorphic processor implementations, achieving orders of magnitude improvements in energy efficiency and computational latency over traditional deep learning approaches.
Comparable algorithmic performance was recently made possible as well with the adaptation of supervised training algorithms to the context of spiking neural networks.
However, information including audio, video, and other sensor-derived data are typically encoded as real-valued signals that are not well-suited to spiking neural networks, preventing the network from leveraging spike timing information.
Efficient encoding from real-valued signals to spikes is therefore critical and significantly impacts the performance of the overall system.
To efficiently encode signals into spikes, both the preservation of information relevant to the task at hand as well as the density of the encoded spikes must be considered.
In this paper, we study four spike encoding methods in the context of a speaker independent digit classification system: \emph{Send on Delta}, \emph{Time to First Spike}, \emph{Leaky Integrate and Fire Neuron} and \emph{Bens Spiker Algorithm}.
We first show that all encoding methods yield higher classification accuracy using significantly fewer spikes when encoding a bio-inspired cochleagram as opposed to a traditional short-time Fourier transform.
We then show that two \emph{Send On Delta} variants result in classification results comparable with a state of the art deep convolutional neural network baseline, while simultaneously reducing the encoded bit rate.
Finally, we show that several encoding methods result in improved performance over the conventional deep learning baseline in certain cases, further demonstrating the power of spike encoding algorithms in the encoding of real-valued signals and that neuromorphic implementation has the potential to outperform state of the art techniques.
\end{abstract}



\begin{teaserfigure}
  \includegraphics[width=\textwidth]{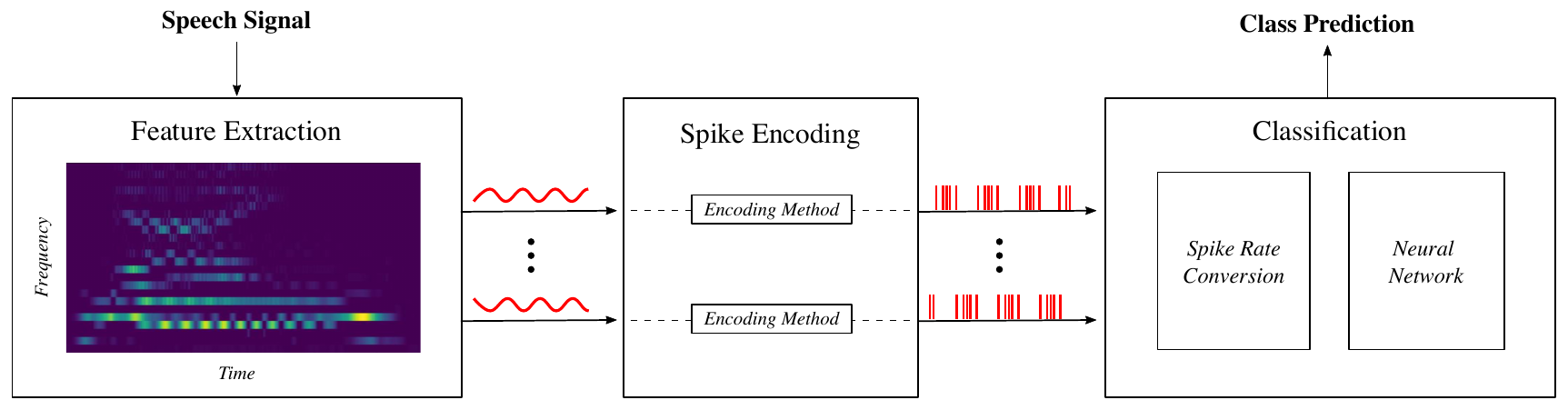}
  \caption{Architecture of the proposed system for evaluation of spike encoding methods. The feature extraction module (left) comprises either a spectrogram or bio-inspired cochleagram. Each frequency channel is then encoded into spikes with the encoding method under study in the Spike Encoding module (center). The encoded spikes are then converted for compatibility with a conventional reference CNN that performs the final classification (right).}
  \Description{Architecture of the proposed system for evaluation of spike encoding methods. The feature extraction module (left) comprises either a spectrogram or bio-inspired cochleagram. Each frequency channel is then encoded into spikes with the encoding method under study in the Spike Encoding module (center). The encoded spikes are then converted for compatibility with a conventional reference CNN that performs the final classification (right).}
  \label{img:Architecture}
\end{teaserfigure}

\maketitle

\section{Introduction}
Spiking Neural Networks (SNN) encode information in an event-driven fashion using spikes that are dynamically transmitted between neurons in the network. 
Recent work in neuromorphic computing has demonstrated that SNNs can result in significantly lower power requirements compared to traditional state of the art deep learning-based approaches \cite{Davies:18}.
However, signals typically processed using neural networks including audio and video are encoded and stored as real-valued signals that are incompatible with SNNs.
It is therefore necessary to encode these signals as spike trains to enable subsequent processing with an SNN.
This conversion process, referred to as spike encoding, is therefore crucial and has an impact on the overall performance of the system. 

A variety of spike encoding methods have been proposed in the literature covering a range of applications in image and signal processing~\cite{Miskowicz:06, Guo:07, Schrauwen:03}.
For example, Guo \textit{et al.}~\cite{guo2021neural} compare the \emph{Time to First Spike} (TTFS), \emph{Phase}, and \emph{Burst} spike encoding methods in the context of MNIST handwritten digits classification.
In the image classification setting, Kheradpisheh \emph{et al.}~\cite{kheradpisheh2020temporal} propose a backpropagation algorithm to train an SNN on the Caltech face/motorbike and MNIST datasets using a TTFS encoding.
In the context of biosignal processing, Garg \textit{et al.}~\cite{garg2021signals} study the impact of using \emph{Send on Delta} (SOD) encoding for classification of Electromyography (EMG) signals.
In Petro \textit{et al.}~\cite{Petro:20}, synthetic signals are encoded into spikes and then decoded, with the reconstruction quality evaluated for \emph{Bens Spiker Algorithm} (BSA) and compared with 3 other temporal contrast encodings.
To our knowledge, few authors have compared spike encoding methods for speech. The work by Pan \textit{et al.}~\cite{Pan:19,Pan:20} studies spike encodings for speech recognition using the TIDIGITS dataset~\cite{Leonard:93}. They compare \emph{Phase}, \emph{Latency}, and \emph{Threshold} encoding methods and report that \emph{Threshold} encoding provides the highest classification accuracy.

The choice of optimal spike encoding method is dependent on the problem setting, as reported previously in~\cite{Petro:20}.
The complexity of the encoding algorithm and any required parameter optimization need to be taken into account in selecting an encoding method as well.
Furthermore, spike encoding methods have been shown to provide significant data compression, for example as demonstrated in \cite{Sengupta:17} in the context of functional magnetic resonance imaging (fMRI) cognitive state discrimination task.
Decreasing the density of encoded spikes results in reduced activity in the subsequent SNN processing, lowering its energy consumption and further magnifying the gains in energy efficiency provided by SNNs.
Therefore, the resulting encoded spike density is an important consideration in selecting an optimal encoding method as well.

In this work, we aim to gain insights into the impact of the choice of spike encoding method and their resulting spike densities in the context of a speaker independent digit classification task.
We study four popular spike encoding methods in terms of \textit{i)} the generated spike densities, and \textit{ii)} the resulting classification accuracy.
By varying the parameters of the encoding methods, we are able to determine optimal points of operation that simultaneously maximize classification accuracy while minimizing the number of generated spikes.
To our knowledge, this kind of work has not yet been reported in the literature.
We demonstrate that it is possible to reach state of the art classification accuracies in combination with a reduced spike density for speech classification using this approach.
Finally, we provide an analysis of the impact of the parameters of each encoding method that affect the resulting spike density.

\section{System Architecture}\label{sec:system-architecture}

A block diagram of the overall system architecture is shown in Fig.~\ref{img:Architecture}. The system comprises 3 modules: a feature extraction module (\ref{subsec:Feature extraction}), a spike encoding module (\ref{subsec:Encoding}), and classification  module (\ref{subsec:Classification}). We proceed to describe each of these modules in detail below.

\subsection{Feature Extraction}
\label{subsec:Feature extraction}
The first step of the system is to extract relevant features from the speech signals. We consider two time-frequency transforms here, including a Fourier transform-based spectrogram and a bio-inspired cochleagram allowing us to explore the impact of auditory system-inspired time-frequency processing and its affect on the various spike encoding methods.

\subsubsection{Spectrogram}
The spectrogram feature extraction method consists of a short-time Fourier transform (STFT) with a sliding 5~ms Tukey window and a 0.5~ms frame advance. The 5~ms window size was chosen to capture speech characteristics relevant in the context of a speech recognition task. We retain the first 24 frequency points of the STFT transform, such that the resulting spectrogram covers the frequency range from 0~Hz to 4600~Hz.

\subsubsection{Cochleagram}

The cochleagram feature extraction method is based on an efficient bio-inspired model of the time-frequency processing performed by the auditory system \cite{Adeli:16}, with source code provided online\footnote{https://github.com/NECOTIS/Adeli-Timbre-Hierarchical-Model}.
Due to the speech recognition task we study here, the time averaging step this model applies is not used in our experiments as it leads to loss of relevant information.

The cochleagram comprises a bank of 24 cochlear filters with center frequencies ranging from 100~Hz to 4500~Hz. For each frequency channel, the envelope is then computed followed by a downsampling operation by a factor of 10.
Compression is then applied by taking the square root of the downsampled envelopes.
Finally, the last processing step involves performing lateral inhibition between frequency channels, followed by half-wave rectification.

After the time-frequency transforms are computed, both the spectrogram and cochleagram are downsampled by a factor of 2 yielding an equivalent sampling frequency of 1000~Hz. In the final time-frequency representation, features extracted for each input signal are encoded using 24 frequency points with the number of windows equal to the number of samples of the input signal divided by 20.

\subsection{Spike Encoding}
\label{subsec:Encoding}
Prior to being converted into spikes, the time-frequency features are normalized between 0 and 1.
For a given utterance, an amplitude of 1 is attributed to the sample that has the maximum amplitude in the overall representation across all frequency channels and time windows.
Each frequency channel is then encoded using the spike encoding method under study, where the various encoding methods are presented in detail in Section \ref{sec:encoding-methods}.
After the spike conversion process, spike trains are encoded via the Address Event Representation (AER) protocol \cite{Jimenez:09}, where each spike is represented using a pair of values consisting of the corresponding channel number and the time instant of its occurrence.

\subsection{Classification}
\label{subsec:Classification}
In the context of spoken digit classification, Convolutional Neural Networks (CNN) have been shown to yield very high classification accuracy when compared to other classification methods \cite{Anumula:18}.
Therefore, we chose this classifier as our reference system.
We reproduced the architecture presented in~\cite{Neil:16}, consisting of 4 convolution layers each comprising 32 convolution filters of size 3x3 followed by a Rectified Linear Unit (ReLU) activation function and Average Pooling. The architecture also contains dense and dropout layers.

To perform classification with a conventional CNN, encoded spike trains need to be converted into real-valued signals. This is done for each frequency channel using a finite impulse response (FIR) averaging filter with a 5~ms impulse response. As the reference CNN was initially designed for the Dynamic Audio Sensor (DAS)~\cite{Chan:07} which uses 64 channels, we zero-pad the decoded spectrograms and cochleagrams in time and frequency/channels to reach the same duration and number of channels compatible with the reference CNN.

\section{Encoding Methods}\label{sec:encoding-methods}
\label{sec:coding methods}
In this section, we present the spike encoding methods that encode the time-frequency representations presented above into spikes, namely \emph{Send on Delta} (SOD), \emph{Time to First Spike} (TTFS), \emph{Leaky Integrate and Fire Neuron} (LIF) and \emph{Bens Spiker Algorithm} (BSA).

\subsection{Send on Delta}
\begin{figure}[t]
\centerline{\includegraphics[scale=1.0]{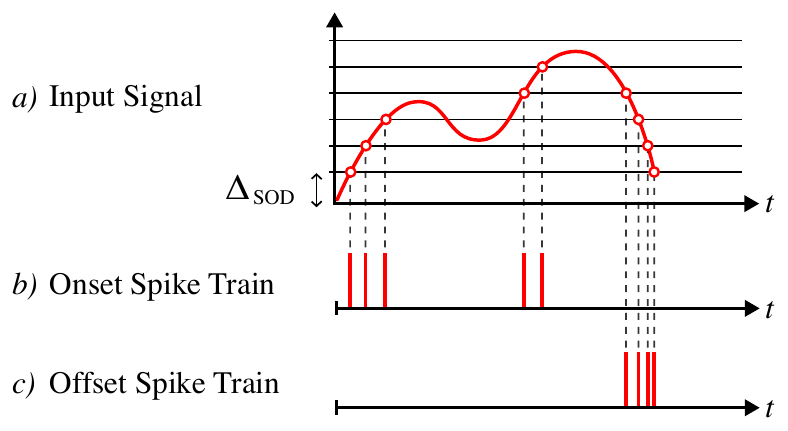}}
\caption{ SOD encoding method. The instants when the variation of the signal's amplitude is greater than or equal to a predefined threshold $\Delta_\textrm{SOD}$ are marked. An upward variation corresponds to an onset, while a downward variation corresponds to an offset, resulting in the SOD\textsubscript{ON} and SOD\textsubscript{OFF} spike trains respectively.}
\Description{SOD encoding method. The instants when the variation of the signal's amplitude is greater than or equal to a predefined threshold $\Delta_\textrm{SOD}$ are marked. An upward variation corresponds to an onset, while a downward variation corresponds to an offset, resulting in the SOD\textsubscript{ON} and SOD\textsubscript{OFF} spike trains respectively.}
\label{img:SOD_illustration}
\end{figure}

The \emph{Send on Delta} (SOD) spike encoding method encodes significant amplitude variations in either positive or negative directions as spikes \cite{Miskowicz:06}.
We present the SOD algorithm in Alg. \ref{alg:sod} and illustrate it schematically in Fig.~\ref{img:SOD_illustration}.
The encoding process consists of iterating over the input signal sample by sample and generating a spike when a significant amplitude variation is observed.
The variation is computed by taking the difference between the current signal value and the amplitude at which the previous spike was generated.
A variation is considered significant when its absolute value is greater than or equal to a predetermined threshold $\Delta_\textrm{SOD}$.
The spikes generated for positive variations (increase) are stored separately from those generated for negative variations (decrease), resulting in 48 spike trains given a time-frequency representation of 24 frequency points.
We study two additional SOD variants in our experiments for which we retain only the signal increases with SOD\textsubscript{ON}, or the signal decreases with SOD\textsubscript{OFF}.

\begin{algorithm}[h]
\caption{SOD}
\label{alg:sod}
\begin{algorithmic}
\STATE $t \gets 0$
\STATE $t\textsubscript{ref} \gets 0$
\WHILE{$t < N$}
\IF{$y[t]-y[t\textsubscript{ref}] \geq \Delta_\textrm{SOD}$}
    \STATE $SOD\textsubscript{ON} \gets t$
    \STATE $t\textsubscript{ref} \gets t$
\ELSIF{$y[t\textsubscript{ref}]-y[t] \geq \Delta_\textrm{SOD}$}
    \STATE $SOD\textsubscript{OFF} \gets t$
    \STATE $t\textsubscript{ref} \gets t$
\ENDIF
\STATE $t \gets t+1$
\ENDWHILE
\end{algorithmic}
\end{algorithm}

\subsection{Time to First Spike}
\begin{figure}[ht]
\centerline{\includegraphics[scale=1.0]{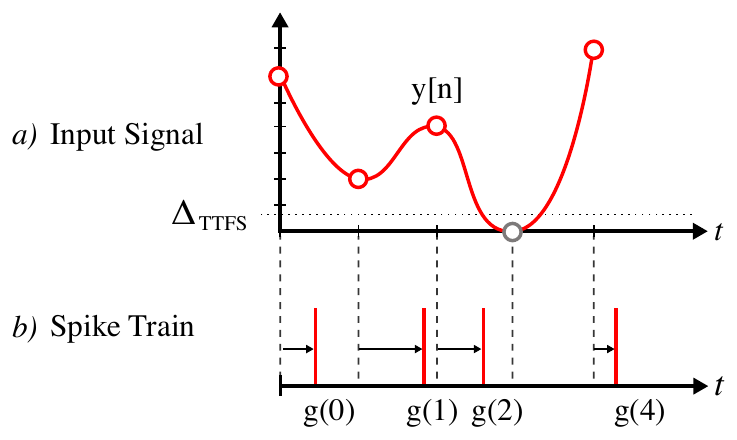}}
\caption{ TTFS encoding method. A spike is generated for all samples with values greater than $\Delta_\textsubscript{TTFS}$. Larger amplitude values result in spikes generated sooner, while smaller amplitudes result in spikes generated later. For example, the corresponding spike for $y[0]$ occurs at time g(0), while no spike is generated for $y[2]$ because it is less than $\Delta_\textsubscript{TTFS}$.}
\Description{ TTFS encoding method. A spike is generated for all samples with values greater than $\Delta_\textsubscript{TTFS}$. Larger amplitude values result in spikes generated sooner, while smaller amplitudes result in spikes generated later. For example, the corresponding spike for $y[0]$ occurs at time g(0), while no spike is generated for $y[2]$ because it is less than $\Delta_\textsubscript{TTFS}$.}
\label{img:TTFS_illustration}
\end{figure}

The \emph{Time to First Spike} (TTFS) spike encoding method is a time-based method typically used to encode images \cite{Guo:07, guo:21}.
In the context of image processing, TTFS encodes each pixel as a spike that occurs sooner or later depending on the pixel's value.
High intensity pixels are encoded with spikes that arrive earlier while low intensity pixels are represented by spikes that arrive later. 
In this work, we derive a modified TTFS encoding that does not encode small amplitude samples and comprises a logarithmic scale. 
Samples $y[n]$ for which $y[n]<\Delta_\textrm{TTFS}$ are ignored (see Fig. \ref{img:TTFS_illustration}). 
For a given sample index $n$, we define the function $g(n)$ in Eq.~\eqref{equation:TTFS} that is equal to the shifted time instant associated with the sample $y[n]$,
\begin{equation}
g(n) = \left(n + \frac{\log(y[n])}{\log(\Delta_\textrm{TTFS})}\right) T_s \label{equation:TTFS}
\end{equation}
where $T_{s}=1/f_s$ is the sampling period where the sampling frequency $f_s$ after extraction of the time-frequency representations is 1000 Hz in our experiments.
We note that $g(n)$ encodes the exact instant when the associated spike occurs.
Therefore, unlike the other spike encoding methods presented here, the instant at which the spike occurs is continuous.

\subsection{Leaky Integrated and Fire}
\begin{figure}[htbp]
\centerline{\includegraphics[scale=1.0]{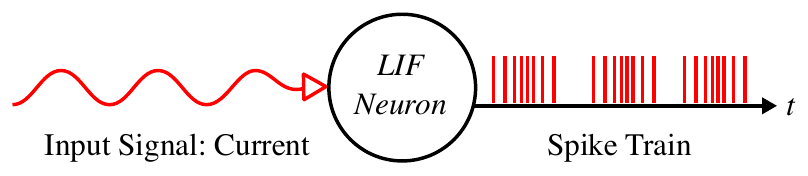}}
\caption{ LIF encoding method. The input signal is introduced into the neuron as current, which leads to generation of spikes according to its differential equation defined in Eq. \eqref{equation:LIF}.}
\Description{ LIF encoding method. The input signal is introduced into the neuron as current, which leads to generation of spikes according to its differential equation defined in Eq. \eqref{equation:LIF}.}
\label{img:LIF_illustration}
\end{figure}

The \emph{Leaky Integrated and Fire} (LIF) neuron method is a commonly-used approach to encode real-valued signals into spike trains. This method is biologically plausible.
For a given frequency channel index $i$, the signal $y_i[n]$ is provided as input current to an LIF neuron with the same index.
Spikes are generated when the neuron's potential reaches a pre-determined threshold $\Delta_\textrm{LIF}$.
This process is illustrated in Fig.~\ref{img:LIF_illustration}.
We note that each LIF neuron $i$ has its own time constant $\tau_{i}$, while the spike threshold $\Delta_\textrm{LIF}$ is the same for all neurons.

The differential equation of neuron $i$ is defined as,
\begin{equation}
\frac{dV_i}{dt} = \frac{I_i-V_i}{\tau_i} \label{equation:LIF}
\end{equation}
where $V_i$ represents the neuron's membrane potential and $I_i$ represents its current, i.e. $I_i=y_i[n]$. We note that this is a simplified version of the standard LIF neuron where the resting potential is set to 0 V and the membrane resistance is set to 1~$\Omega$.  

\subsection{Bens Spiker Algorithm}
\emph{Bens Spiker Algorithm} (BSA) is widely used in time series encoding \cite{Schrauwen:03, Nuntalid:11, Sengupta:15} and belongs to the family of so-called stimulus estimation methods.
These methods are based on the principle that the stimulus of a biological neuron can be estimated from a sequence of spikes by filtering it linearly.
BSA involves convolving a finite impulse response (FIR) filter with the input signal and generating a spike when the difference between the filtered signal and the input signal is less than a defined threshold.
An important challenge in using this method therefore lies in the choice of both the filter parameters and the threshold value.
The typical approach is to perform a grid search by using the Signal to Noise Ratio (SNR) as error metric \cite{Petro:20}.

\section{Experiments}\label{sec:experiments}

\subsection{Dataset}
We use the TIDIGITS dataset \cite{Leonard:93} to compare the spike encoding methods presented above.
TIDIGITS is a collection of spoken utterances consisting of 11 classes, 0-9 and ``oh'', sampled at 20 kHz.
While the dataset includes utterances from children and adults, we focus on single digit utterances pronounced by adult speakers, representing 2464 training examples and 2486 test examples.
For comparison purposes, we also use the neuromorphic N-TIDIGITS dataset \cite{Liu:18} that was created by pre-encoding the same subsets of TIDIGITS as above using the Dynamic Audio Sensor (DAS) \cite{Liu:14}.

\subsection{Metrics}
The metrics we use for evaluation of the spike encoding methods consist of the classification accuracy and the encoded spike density.
Classification accuracy represents the percentage of examples in the test set that were correctly classified.
The spike density is defined for a given speech signal as the ratio between the number of generated spikes and the number of samples $y[n]$ of the channels of the spectro-temporal representations, i.e. the spectrogram or cochleagram, just before spike encoding.
The spike density is then averaged over all examples in test set.
In the experiments below, we compare the classification accuracy of each method over a range of different spike densities, where the aim is to maximize classification performance while simultaneously maximizing energy efficiency by minimizing spike density. So we assume that low spike density leads to low energy consumption.

\begin{figure*}[htbp]
\centerline{\includegraphics[scale=0.95]{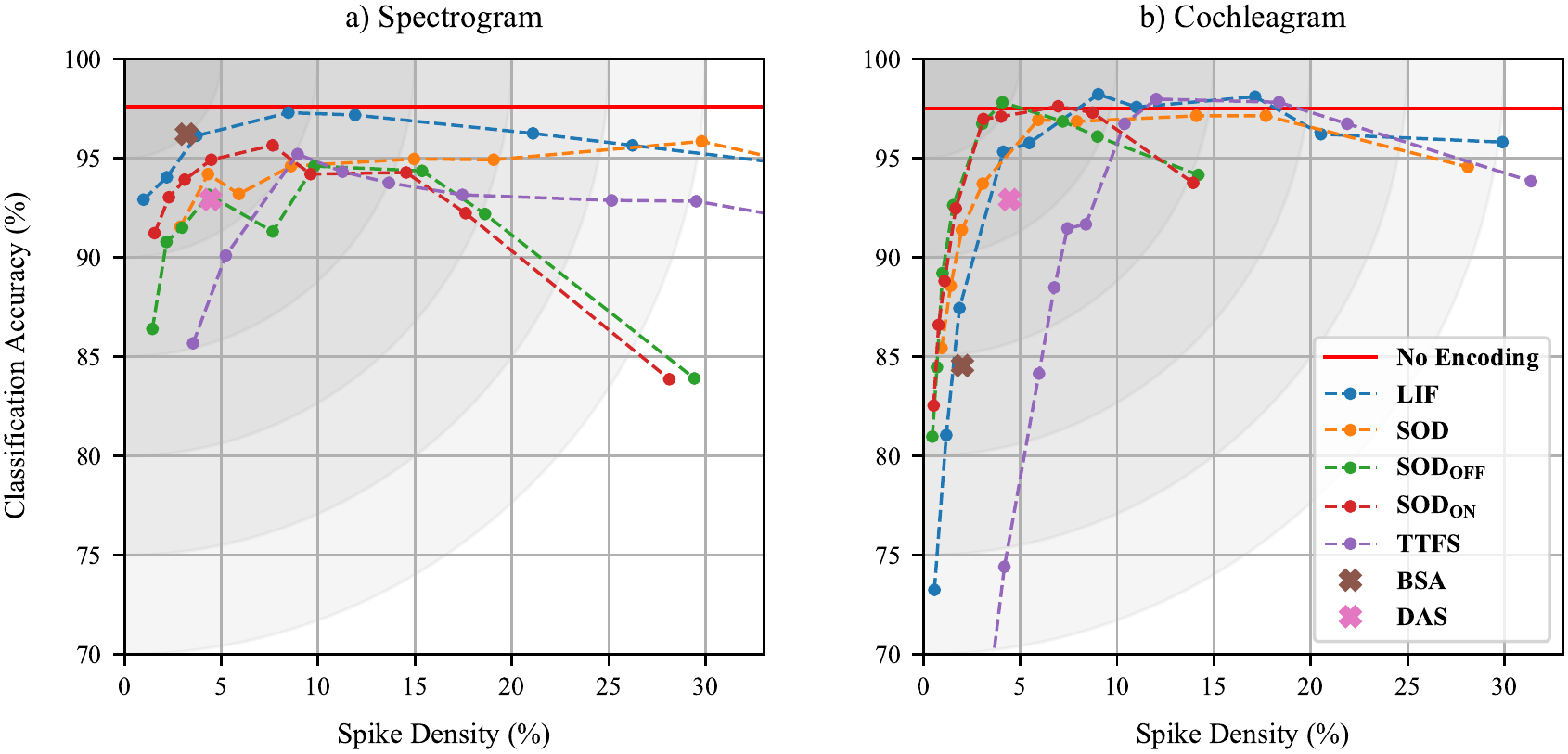}}
\caption{ Variation of classification accuracy according to spike density. The \emph{No Encoding} label defines the classification accuracy which is obtained without encoding. The DAS approach includes its own feature extractor and encoding method and is cited here for comparison purposes.}
\Description{ Variation of classification accuracy according to spike density. The \emph{No Encoding} label defines the classification accuracy which is obtained without encoding. The DAS approach includes its own feature extractor and encoding method and is cited here for comparison purposes.}
\label{img:results}
\end{figure*}

\subsection{Experimental Process}
Each spike encoding method has parameters that control the resulting encoded spike density.
We aim to determine the relationship between spike density and resulting classification accuracy for each of the encoding methods.
Preliminary exploration suggested that spike densities less than 30\% were sufficient to capture a wide range of classification accuracies for all encoding methods, with the maximum for each method falling well within this range.
Each method’s parameters are chosen as follows to uniformly cover the desired range of encoded spike densities of 0 to 30\%.:
\begin{itemize}
  \item \textbf{SOD}: The threshold parameter $\Delta_\textrm{SOD}$ values are chosen such that they cover the interval [$10^{-1}$,  $10^{-4}$].
  \item \textbf{TTFS}: The threshold parameter $\Delta_\textrm{TTFS}$ values are chosen to cover the interval [$3.10^{-1}$, $10^{-4}$].
  \item \textbf{LIF}: Time constants $\tau_i$ are inversely proportional to the center frequency of channel $i$. They were empirically chosen in the range [20, 40]~ms with the tradeoff of achieving good classification accuracy with the fewest number of spikes. The threshold parameter $\Delta_\textrm{LIF}$ is the same for all channels and covers the interval [$10^{-1}$, $10^{-5}$].
  \item \textbf{BSA}: We calculated the optimal filters for BSA using 10\% of the training set, based on minimizing the SNR.
\end{itemize}

In addition to these experiments, a \emph{No Encoding} classification accuracy was calculated by performing the experiment directly on the time-frequency representations (spectrogram or cochleagram), thus bypassing the spike encoding stage shown in Fig. \ref{img:Architecture}.
The network is therefore trained and evaluated on the output of the raw spectrogram or cochleagram outputs.
This allows us to evaluate the potential loss of information required for classification due to the spike encoding methods.
For all experiments, the CNN was trained over 50 epochs using the Adam optimizer with a weight decay of 0.0, a learning rate of 0.001, and a batch size of 8.

\section{Results}\label{sec:results}
In Fig.~\ref{img:results}, we present the experimental results demonstrating the effect on classification accuracy for the different spike encoding methods as the spike density is varied as described above.
The results are shown for the two feature extraction methods presented in \ref{subsec:Feature extraction}, namely the spectrogram and the bio-inspired cochleagram.
We note that for all encoding methods and for both feature extraction methods, the classification accuracies follow the same general trend as the spike density is varied from low to high.
As the encoded spike density increases, we observe three major stages: a rapid growth, a stagnation, and a decay.

Comparing the results for the spectrogram (Fig.~\ref{img:results}a) and the cochleagram (Fig.~\ref{img:results}b), we see that the encoding algorithms tend to generate notably more spikes for the spectrogram than for the cochleagram.
This trend is evidenced by the curves being shifted to the right for the spectrogram when compared to the cochleagram.
This is likely due to the cochleagram's lateral inhibition module that allows competition between neighboring channels \cite{Adeli:16}, resulting in the suppression of channels with low energy.

\begin{figure}[t]
\centerline{\includegraphics[]{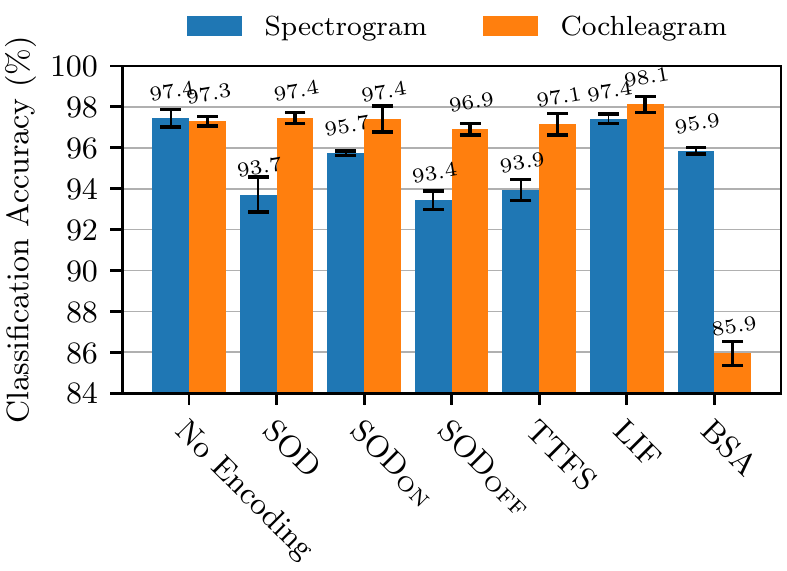}}
\caption{ Best classification accuracy for each encoding method}
\Description{ Best classification accuracy for each encoding method}
\label{img:best_res}
\end{figure}

For each method, the experiments for which the parameters gave the best classification accuracies in Fig.~\ref{img:results} were repeated 6 times by varying the random seed used to train the CNN in order to estimate the mean and standard deviation.
The results are reported with corresponding spike densities in Table~\ref{tab:results}.
The classification accuracy comparison between these methods are presented in Fig.~\ref{img:best_res}.
The \emph{No Encoding} classification accuracy of the cochleagram (97.3\%) is approximately equal to that of the spectrogram (97.4\%).
However, with the cochleagram there are more methods that achieve the \emph{No Encoding} classification accuracy.
Moreover, with the cochleagram the variants of SOD reach the first quarter of a circle (Fig.~\ref{img:results}b), corresponding to less than 5\% of classification errors using less than 5\% of spike density.
Finally, with the exception of BSA, all the methods yield a higher maximum classification accuracy with the cochleagram than with the spectrogram (Fig.~\ref{img:best_res}).

The BSA method performs better with the spectrogram than the cochleagram by almost 10\% (95.86\% vs. 85.95\%).
This can likely be explained by the fact that in the cochleagram, the frequency scale is nonlinear \cite{Adeli:16}.
In fact, in the bank of cochlear filters used, the high frequency filters have larger bandwidths.
As a result, the filters used by BSA to estimate the high frequency stimuli are less precise.
All methods achieve maximum classification accuracy with less than 30\% spike density.
Also the studied methods result in better maximum classification accuracy than that of the DAS N-TIDIGITS approach.
The SOD\textsubscript{ON} and SOD\textsubscript{OFF} variants offer a good compromise, yielding similar classification accuracies to SOD with almost 2 times fewer spikes.
In addition, SOD and LIF achieve higher classification performance using cochleagram feature extractor than the previously reported state of the art (97.4\%) which uses a Biologically plausible Auditory Encoding (BAE) as encoding scheme and an SNN classifier with Membrane Potential Driven Aggregate-Label Learning (MPD-AL) as learning rule \cite{Pan:20}.

\begin{table}[]\centering
\setlength{\tabcolsep}{2pt}\caption{Best classification accuracy and corresponding spike density obtained for each spike encoding method}
\begin{tabular}{ l @{\hskip 6pt} c c c c @{\hskip 8pt} c c c }
&& \multicolumn{2}{c}{Spectrogram} & \phantom{} & \multicolumn{2}{c}{Cochleagram} \\[1pt] 
\cmidrule{2-4} \cmidrule{6-7}
&& Accuracy & Spike Density && Accuracy & Spike Density \\
&& (\%) & (\%) && (\%) & (\%) \\ \midrule
No Encoding & &  \textbf{97.44} & - && 97.30 & - \\ \addlinespace[3pt]
SOD & &  93.71 & 29.81 && \textbf{97.45} & \textbf{14.09} \\
SOD\textsubscript{ON} &&  95.74 & 07.65 && \textbf{97.40} & \textbf{06.96} \\
SOD\textsubscript{OFF} &&  93.44 & 09.80 && \textbf{96.90} & \textbf{04.08} \\ \addlinespace[2pt]
TTFS & &  93.95 & 08.94 && \textbf{97.14} & \textbf{12.02} \\ \addlinespace[2pt]
LIF & &  97.42 & 08.46 && \textbf{98.12} & \textbf{09.03} \\ \addlinespace[2pt] 
BSA & & \textbf{95.86} & \textbf{03.20} && 85.95 & 02.02 \\
\bottomrule
\end{tabular}
\label{tab:results}
\end{table}

Finally, Table \ref{tab:literature} reports a comparison of classification results with previous TIDIGITS speech recognition approaches from the literature.
We note that while results achieved in this work are generally comparable to those in the literature, the LIF method we present yields the highest classification accuracy reported thus far of 98.1\%.

\section{Discussion}\label{sec:discussion}
\subsection{Cochleagram, $SOD\textsubscript{ON}$ \& $SOD\textsubscript{OFF}$, LIF, Curves trend}
Cochleagrams shift upward most classification accuracies in comparison with the Fourier-based spectrogram. Only the reference system (\emph{No Encoding}) has the same accuracies for either the spectrogram or cochleagram. Cochleagrams allow a spike density reduction for all methods. Moreover, it allows the encoding methods (except BSA) to reach the \emph{No Encoding} classification accuracy (97\%), suggesting that all relevant information required for classification with the CNN is preserved after spike encoding of the cochleagram.

With the cochleagram, SOD\textsubscript{ON} and SOD\textsubscript{OFF} are the most efficient encodings, providing conservation of relevant information for classification with the CNN and with a reduced spike density. While achieving the \emph{No Encoding} classification accuracy (97\%), they only use 7\% and 4\% spike densities respectively.

LIF was found to be the most robust method, as it succeeds in achieving the \emph{No Encoding} classification accuracy for both the spectrogram and cochleagram representations.
It also achieves the best classification accuracy over all experiments (98.12\%).

In Fig.~\ref{img:results}, it is remarkable that all the curves follow the same trend as mentioned in the previous section.
The general trend can be divided into three major stages and our hypotheses are as follows,
\begin{itemize}
  \item Rapid growth: the spike encoding methods succeed in encoding increasing amounts of relevant information and the classifcation rate increases rapidly.
  \item Stagnation: the information necessary for classification has already been encoded, and since no useful information is added, the classification accuracy no longer increases.
  \item Decay: the encoding methods encode information not relevant to the classification which therefore acts like noise thus leading to a decrease in classification accuracy.
\end{itemize}

\begin{table}[h]
\caption{TIDIGITS Classification Accuracy Comparison Between Proposed Method and Existing Methods in the Literature }
\begin{center}
\begin{tabular}{ l c } 
\textbf{Method} & \textbf{Accuracy (\%)}  \\ 
\midrule
 Cochleagram - LIF - CNN (this work)   & \textbf{98.1} \\ 
 Cochleagram - SOD - CNN (this work)   & 97.4 \\ 
 Cochleagram - TTFS - CNN (this work)   & 97.1 \\ 
 \addlinespace[2pt]
 BAE - MPDAL (Pan, Zihan, et al. \cite{Pan:20})  & 97.4 \\ 
 AER silicon cochlea - SVM (Abdollahi \& Liu \cite{abdollahi:11})  & 95.6 \\
 AMS1c - GRU RNN (Anumula, Jithendar, et al.  \cite{Anumula:18})  & 91.1 \\
\bottomrule
\end{tabular}
\label{tab:literature}
\end{center}
\end{table}

\subsection{Bit Compression is Feasible}
In general, spike encoding makes it possible to reduce the volume of information while preserving information relevant for classification.
This allows the encoding methods to provide a significant reduction in terms of bit rate.
It would be possible to estimate the bit compression ratio (BCR) as defined in \cite{Sengupta:17} for each encoding method depending on the processor and hardware to be used.
For the TIDIGITS database, for example, the sampling frequency $f_s$ is 20 kHz and samples are encoded with a bit depth of 32 bits, resulting in an initial bit rate of 640 kbps.
With the configuration that we have in this setup (channel outputs at 1 kHz, with 24 channels and a 16 bit AER representation of spikes), BCR is on the order of 0.06 for an average spike density of 10\%.
Note that 10\% spike density is reasonable, as the best performance was obtained with smaller spike density.
In the context of speech recognition, however, it is common to use features based on MFCC coefficients and their delta representations.
For example, given an 32-dimensional MFCC vector transmitted every 5 ms, with real values encoded with a bit depth of 32 bits, our resulting BCR would instead be on the order of 0.19.

\subsection{Specific Encoding Method Characteristics and Constraints}
Each spike encoding method has characteristics and constraints that are important to consider when interpreting the results and selecting a desired method.

\begin{itemize}
  \item SOD has a single parameter which makes it simple to optimize. Its algorithm is also simple and straightforward to implement.
  However, it could be improved by using a different threshold per channel.
  The encoded information is dependent on the variation of the gross envelope of the signal.
  It would not be possible to easily and exactly reconstruct the signal, but it is sufficient to retain features for classification with a CNN.
  \item LIF has two parameters which makes its optimization more difficult. However, it is biologically plausible and easy to implement.
The encoded information is related to a short-term integration of the signal. It would not be possible to easily and exactly reconstruct the signal, but it is also sufficient to retain features for classification with a CNN.

  \item TTFS has one parameter and is therefore easy to optimize. Its implementation is also simple.
The encoded information is a quasi-direct estimation of the log signal (a part from the fact that we use a small threshold). It would therefore be possible to easily and quasi exactly reconstruct the signal. But the classifier which is used here (Spike decoding + CNN) is not able to exploit the timing of the spikes very well. With the use of the cochleagram, TTFS is able to beat the reference system (but with the cost of a higher spike density). This is not the case with the spectrogram which removes too much of the time encoded features of speech. We suspect that the classifier we used (Spike decoding + CNN) may be biasing the results here.

\item 
BSA is more complex than the other methods and more difficult to implement. It has three parameters, making it difficult to optimize as well. However, by using a subset of the dataset, we can find the optimal parameters before carrying out classification. As was the case with TTFS, the classifier (Spike decoding + CNN) potentially biases the results for BSA as well. BSA was designed for signal encoding \emph{and} decoding with minimal reconstruction error. 
\end{itemize}
Given that our experimental setup was designed to compare spike encoding methods for speech recognition, we note that SOD and LIF are \emph{destructive} in the sense that they provide very high compression at the expense of not being able to reconstruct the input signal\footnote{It would be possible to reconstruct the original signal, however this would require a population of neurons/filters instead of a single one.}. Despite reducing the amount of information, they preserve and even enhance features meaningful for speech recognition. As such, they represent features that are potentially useful not only for SNNs, but for more conventional deep neural networks including CNNs as well.

\subsection{Dependence on datasets, on the classifier and future work}
We provided a comparison of spike encoding algorithms using the standard TIDIGTS speech classification dataset consisting of clean, isolated speech signals. We therefore cannot guarantee that the results are generalizable to more complex datasets.
Future work will involve a detailed analysis of the impact of additive and convolutive noise on the resulting classification accuracy of the various spike encoding methods.
We will also analyze the energy consumption of the classification phase in order to thoroughly assess the impact of the encoding step in the overall classification system.
Finally, we will replace the classification with an SNN, resulting in an end-to-end spiking solution suitable for integration on a neuromorphic hardware platform. This will remove some of the biases that are commonly observed when comparing spike encoding methods based on a specific application and with the use of a non spiking classifier. For example, TTFS should provide better results as the conversion method here loses fine-grained timing.

\section{Conclusion}\label{sec:conclusion}
In this work, we have studied the variation of classification accuracy as a function of spike density for a variety of spike encoding methods in an isolated digit classification system.
We showed that the use of a bio-inspired cochleagram favors spike encoding methods when compared to a more traditional Fourier-based spectrogram.
We also showed that SOD encoding method variants are efficient achieving high classification accuracy (97\%) with less than 7\% spike density. The LIF method was found to be robust as it achieves the \emph{No Encoding} classification accuracy with both feature extractors.
Finally, we demonstrated that all encoding methods can be optimized to achieve interesting classification accuracies with less than 30\% spike density, corresponding to a bit compression ratio (BCR) of approximately 0.18.
By optimizing the LIF method, we then obtained 98.12 \% classification accuracy which compares favorably to the previously reported state of the art (97.4\%) \cite{Pan:20}.

Thanks to their resulting bit compression ratios, the spike encoding methods studied here might be of interest not only to the SNN community but also to the deep learning community. In fact, our results indicate that these encoding methods could be used as promising features for speech recognition in combination with a cochleagram and a more conventional classifier like a CNN. It is important to notice that even if we had to convert back spikes into real values we obtained state of the art classification. Therefore, a complete spiking implementation should have the potential to outperform state of the art techniques.

\begin{acks}
The authors would like to thank FRQNT équipe and NSERC for funding our research. We would also like to thank Ismaël Balafrej and Ahmad El Ferdaoussi for inspiring discussions during the development of this work.
\end{acks}

\bibliographystyle{ACM-Reference-Format}
\bibliography{biblio}

\end{document}